\documentclass{article}

\PassOptionsToPackage{numbers, compress}{natbib}

\usepackage[final]{bdl_2018}
\usepackage[utf8]{inputenc} 
\usepackage[T1]{fontenc}    
\usepackage{hyperref}       
\usepackage{url}            
\usepackage{booktabs}       
\usepackage{amsfonts}       
\usepackage{nicefrac}       
\usepackage{microtype}      
\usepackage[figuresright]{rotating}
\usepackage[english]{babel}
\usepackage{multicol}
\usepackage{enumitem}
\usepackage{fancyhdr}
\usepackage{adjustbox,lipsum}
\usepackage{amsmath,amssymb,bm}
\usepackage{etoolbox}
\usepackage{float}
\usepackage{graphicx}
\usepackage{color}
\usepackage{accents}
\usepackage{mwe}
\usepackage{algorithm,algpseudocode}
\usepackage{subfig}
\usepackage{marginnote}
\usepackage{wrapfig}
\usepackage[font=footnotesize]{caption}
\usepackage[symbol]{footmisc}

\algnewcommand{\To}{\textbf{To }}
\algnewcommand\Input{\item[\textbf{Input:}]}
\algnewcommand\Initialize{\item[\textbf{Initialize:}]}

\title{Applying SVGD to Bayesian Neural Networks for Cyclical Time-Series Prediction and Inference}

\author{
  Xinyu Hu$^{1, 2}$\footnotemark[2]
  \And Paul Szerlip$^{2}$  
  \And Theofanis Karaletsos$^{2}$ 
  \And Rohit Singh$^{2}$
  \AND
  \texttt{xh2194@cumc.columbia.edu}, \texttt{\{pas,theofanis,rohits\}@uber.com}
  \vspace{-0.85cm}\\
  $^{1}$Columbia University, New York, NY \hfill 
  $^{2}$Uber AI Labs, San Francisco, CA \\
}
\begin{document}
\maketitle

\footnotetext[2]{This work was done while the 
  first author was employed at Uber AI Labs, San Francisco, USA.}

\setlength{\abovedisplayskip}{3pt}
\setlength{\belowdisplayskip}{3pt}
\setlength{\belowcaptionskip}{-12.5pt}

\vspace{-2\baselineskip}
\section{Introduction}
\label{sec:intro}
\vspace{-0.5\baselineskip}
Accurate and robust prediction of time-series data has shown meaningful impact in various applications~\citep{langkvist2014review}. For example, at Uber, predicting rider demand accurately benefits supply planning and resource allocation. Inaccurate predictions and confidence miscalibrations in the estimated predictions can lead to suboptimal decision making which may further result in either over or under supply. Ultimately, in an industrial setting such miscalibrations can result in extra cost to the company or to the customers. However, it is challenging to predict quantities like demand accurately due to potentially unknown exogenous variables that cause anomalous patterns and contribute to prediction variability. 
Although, there are many popular and successful recurrent or modified convolutional network models for capturing time dynamics \citep{hochreiter1997long, gehring2017convolutional}, they typically are trained using maximum likelihood and suffer from overconfidence. 
Moreover, such point estimates are typically insufficient to 
quantify prediction variability. 
While there are important successes using frequentist ensembles~\cite{lakshminarayanan2017simple}, the Bayesian framework is a natural choice for modeling the prediction uncertainty and interpreting the estimates. Recently, many attempts have been made to adapt existing Bayesian techniques to model neural networks \citep{gal2016dropout, gal2016theoretically, blundell2015weight, hernandez2015probabilistic, louizos2017multiplicative, karaletsos2018probabilistic}, referred to as Bayesian neural networks (BNNs). 
Variational inference (VI) is often used to approximate the posterior distribution over parameters efficiently \citep{blundell2015weight}.  One particular posterior approximation for BNNs is the Monte Carlo Dropout~\citep{gal2016dropout, li2017dropout} and has been applied to time-series forecasting as well~\citep{zhu2017deep}. However, both accuracy and scalability of VI depend on the particular approximating distribution. In this work, we employ Stein variational gradient descent (SVGD), which is a generalized nonparametric VI algorithm for approximating continuous distributions \citep{liu2016stein}. 
SVGD has the advantage of not requiring knowledge of the explicit form of the posterior distribution and provides a theoretically guaranteed weak convergence of the samples~\citep{liu2017stein}. By assuming independent prior distributions and using the radial basis function (RBF) kernel, SVGD is fast and scalable to large neural networks
and offers an elegant and efficient solution for forecasting quantities like rider demand while also modeling the prediction uncertainty.

We propose a regression-based BNN model to predict spatiotemporal quantities like hourly rider demand with calibrated uncertainties. The main contributions of this paper are (i) A feed-forward deterministic neural network (DetNN) architecture that predicts cyclical time series data with sensitivity to anomalous forecasting events; (ii) A Bayesian framework applying SVGD to train large neural networks for such tasks, capable of producing time series predictions as well as measures of uncertainty surrounding the predictions. Experiments show that the proposed BNN reduces average estimation error by 10\% across 8 U.S. cities compared to a fine-tuned multilayer perceptron (MLP), and 4\% better than the same network architecture trained without SVGD. 

\vspace{-0.75\baselineskip}
\section{Bayesian neural network}
\label{sec:model}
\begin{figure}[ht]
    \vspace{-0.8\baselineskip}
    \centering
    \includegraphics[width=11cm]{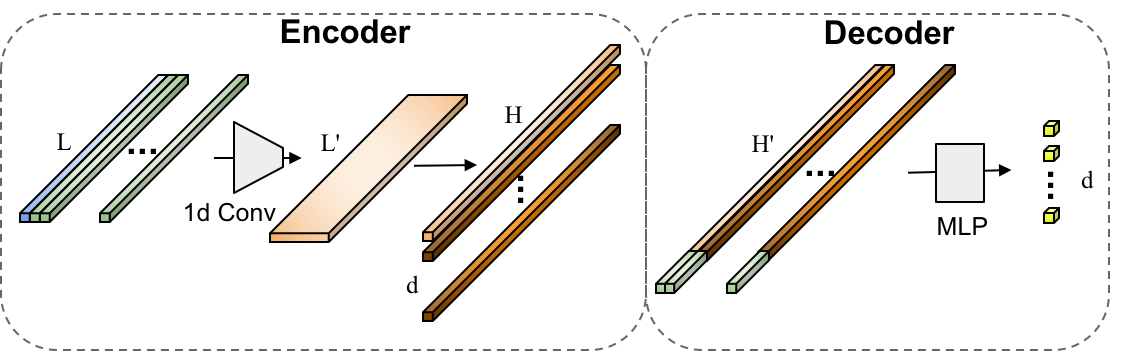}
    \caption{Neural network structure: in the encoder, the blue(leftmost) bar indicates the time series of interest to predict, for example hourly rider demand; the green bars(after the first bar) indicate the sequential location information, for example, one-hot encoded vector indicating if the hour of the day or the day of the week is a special time window like a holiday we need to pay attention to. The inputs are passed into the 1d convolutional layers to learn the hidden features. $d$ parallel linear functions are used to map from the same hidden features to $d$ reconstructed features. In the decoder, those reconstructed features are concatenated with the sequential location information at the prediction hour given the information is known in advance. Then the learned features are passed into a MLP one by one independently to predict the outcome.}
\label{fig:bnn}
\end{figure}

\vspace{-0.5\baselineskip}
The proposed neural network consists of an encoder to learn the hidden features and a decoder to predict time series, as shown in Figure~\ref{fig:bnn}. The outcome of interest is a vector of continuous variables $\bm{y}$. The input features are denoted as $\bm{x}$. The parameter of the model $\bm{\theta}=(\bm{w}, \bm{\Sigma})$ consists of the neural network parameter $\bm{w}$ and the noise covariance matrix $\bm{\Sigma}$. The regression model is specified as: 
\begin{eqnarray}
\label{eq:reg}
    \bm{y}|\bm{\theta} = f_{\bm{w}}(\bm{x}) + \bm{\epsilon},
\end{eqnarray}

where $\bm{\epsilon} \sim N(\bm{0},\bm{\Sigma})$. In equation~\eqref{eq:reg}, $f_{\bm{w}}(\bm{x})$ denotes the output of the neural network. The predicted sequence is modeled independently across time points through the neural network. The correlation among the time points could be modeled through a structured $\bm{\Sigma}$, but in our experiments, $\bm{\Sigma}$ is assumed to be a $d \times d$ positive definite diagonal matrix for simplicity and computational efficiency. The $i$th outcome $\bm{y}_i$, is a vector of length $d$ and is assumed to be independently but not identically sampled from a multivariate Gaussian distribution $N(f_{\bm{w}}(\bm{x}_i), \bm{\Sigma})$ for $i = 1, \dots, N$. 

A Bayesian framework is imposed on the model~\eqref{eq:reg} by assigning prior distributions to the model parameters. The prior of the neural network parameters is given by $\bm{w}|\alpha \sim N(\bm{0},\alpha^{-1}\bm{I})$, $\alpha \sim \Gamma(a_0, b_0)$; 
the prior of the noise covariance
$\bm{\Sigma} = diag(\sigma^2_1, \dots, \sigma^2_d)$, $\sigma^{-2}_i \sim \Gamma(a_1, b_1)$ for $i=1,\dots,d$. $\bm{w}$ and $\bm{\Sigma}$ are esimated to maximzie the joint log-likelihood with different learning rates. 

During training, $n$ such neural networks are built via SVGD. When a new data point is passed into the trained network, $n$ posterior samples of $\bm{\theta}$ are obtained for inference. The predicted outcome is estimated as
$\mathbb{E}(\bm{y}) = \mathbb{E}_{\bm{\theta}}(\mathbb{E}(\bm{y}|\bm{\theta})).$ 
The prediction variability is decomposed
into three sources: model uncertainty, model misspecification and inherent noise. Assuming there is no misspecification, the prediction variability can be estimated through $n$ SVGD samples by
    $Cov(\bm{y}) = Cov_{\bm{\theta}}(\mathbb{E}(\bm{y}|\bm{\theta})) + \mathbb{E}_{\bm{\theta}}(Cov(\bm{y}|\bm{\theta})),$
where $Cov_{\bm{\theta}}(\mathbb{E}(\bm{y}|\bm{\theta}))$ represents the model uncertainty and $\mathbb{E}_{\bm{\theta}}(Cov(\bm{y}|\bm{\theta}))$ represents the inherent noise. Under the assumption of diagonal noise covariance, constructing a credible region is equivalent to constructing a credible interval at each dimension. The $\alpha$-level credible interval is estimated as 
$ 
[\hat{y} - z_{\alpha/2}\hat{\eta}, \hat{y} + z_{\alpha/2}\hat{\eta}],
$
where $z_{\alpha/2}$ is the upper $\alpha/2$ quantile of a standard Gaussian, $\hat{y}=\frac{1}{n}\sum_{i=1}^n f_{\hat{\bm{w}}_i}(x)$, $\hat{\eta} = \sqrt{\frac{1}{n}\sum_{i=1}^n (\hat{\sigma}_i^{2}+ f_{\hat{\bm{w}}_i}(x)^2) - (\frac{1}{n}\sum_{i=1}^nf_{\hat{\bm{w}}_i}(x))^2}$. 

The detailed BNN via SVGD algorithm is shown in the Appendix. In all experiments, an RBF kernel $k(\bm{\theta}_i, \bm{\theta}_j)=exp(-\frac{1}{h}||\bm{\theta}_i- \bm{\theta}_j||_2^2)$ is used with the bandwidth $h=H^2/logn$ where H is the median of the pairwise distances between the SVGD samples. The bandwidth is changed adaptively over iterations. The Stein operator depends on the target posterior only through the score function $\nabla_{\bm{\theta}} \log p(\bm{\theta}|\mathcal{D})= \nabla_{\bm{\theta}} \log p(\bm{\theta},\mathcal{D})$, where $\mathcal{D} = (x_i, y_i)_{i=1}^N$. Thus the exact posterior distribution is not required to be represented explicitly to generate approximate samples from it. To calculate the gradient of $\log p(\bm{\theta}|\mathcal{D})$, we need all the training data. But during training mini-batches of size $b$ are passed into the neural networks. This is fixed by approximating the data likelihood by $\log p(\bm{\theta}, \mathcal{D}) \approx \log p_0(\bm{\theta}) + \frac{N}{b}\sum_{k=1}^b \log p(\mathcal{D}_k|\bm{\theta})$, where $p_0(\bm{\theta})$ is the prior distribution of $\bm{\theta}$.
\vspace{-0.75\baselineskip}
\section{Experiments}
\label{sec:exp}
\begin{figure}[ht]
    \vspace{-1.8\baselineskip}
    \centering
    \subfloat[Overview]{{\includegraphics[width=6cm]{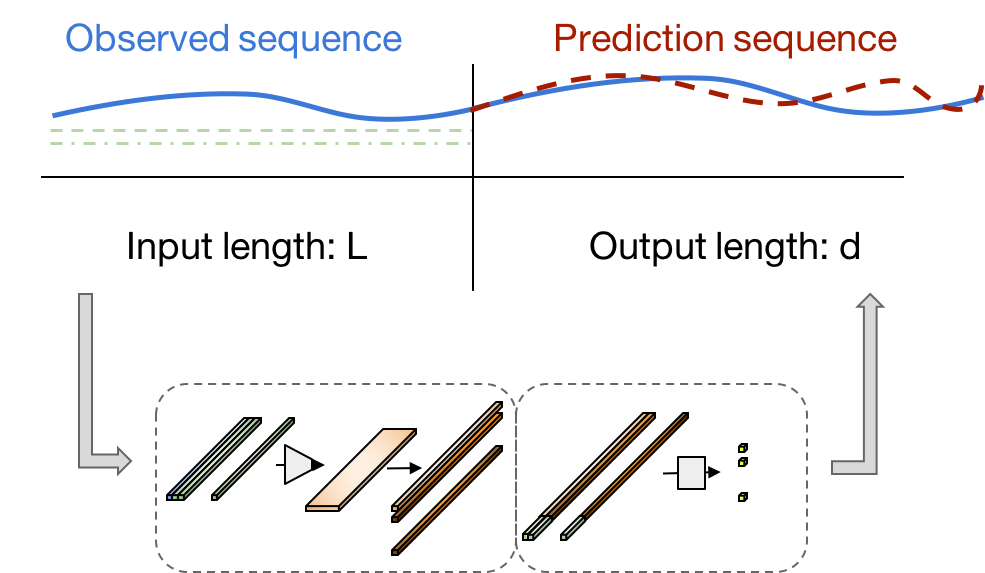} }}
    \qquad
     \vspace{-0.6\baselineskip}
    \subfloat[Estimated prediction variability]{{\includegraphics[width=7cm]{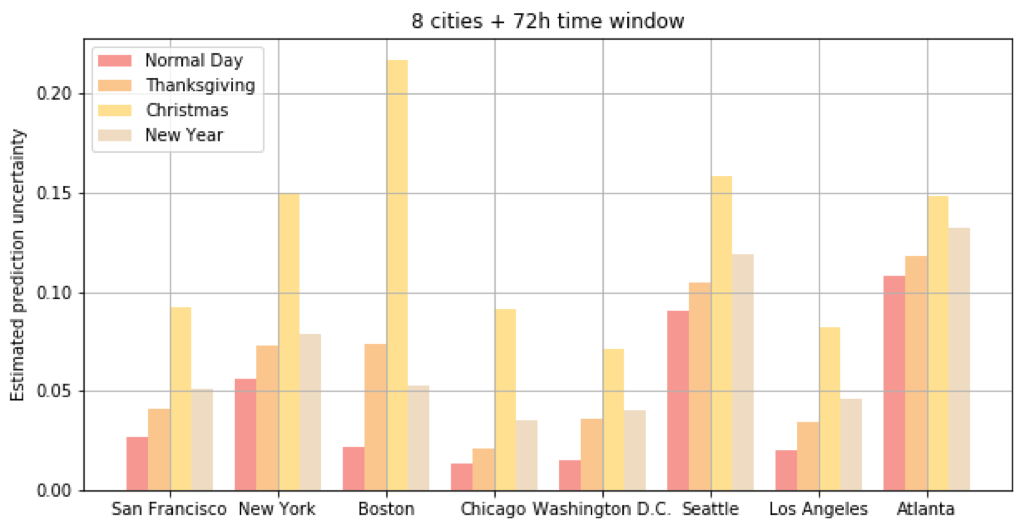}}}
    \caption{ (a) is an overview of the prediction process: the blue line indicates the demand sequence, the green dotted lines indicate the one-hot encoded sequential location input, the red dashed line indicates the predicted sequence; (b) shows the estimated variability when predicting 72 hours around holidays and a non-holiday on the test data in 8 cities using a BNN model with 30 particle samples.}
    \label{fig:overview}
\end{figure}

\vspace{-0.5\baselineskip}
We predict the hourly rider demand across 8 U.S. cities along with quantified prediction variability. The data used in the experiment is the hourly number of completed trips at Uber from 2014 to 2018 among 8 U.S. cities. The dataset is split sequentially into 50\%/25\%/25\% train, validation, and test data and preprocessed to fit the Gaussian assumption. The hourly demand data exhibits a strong 24-hour cyclical pattern with jitters around some special time windows. For example, the demand drops during Thanksgiving and rises dramatically after New Year's eve. To handle the important time windows, extra one-hot encoded channels are added to the input to the convolutional layers. As illustrated in Figure~\ref{fig:overview} (a), the input of the model consists of an hourly demand sequence and several sequential location sequences indicating the hour of the day, day of the week etc., the output is the predicted demand sequence. The difference of the prediction variability of a 72-hour window around holidays and a non-holiday using the previous 144-hour input is shown in Figure~\ref{fig:overview} (b). The estimated variability is always higher around holidays, especially around Christmas, compared to the one around a normal day in all 8 cities, meaning that the BNN model is less confident about predicting a holiday than predicting a non-holiday, as expected. 

\begin{table}[ht]
  \vspace{-1\baselineskip}
  \caption{\footnotesize WMAPE comparison for MLP, DetNN and BNN.}
  \label{tbl:wmape}
  \centering
  {\scriptsize
  \begin{tabular}{lccccc}
	\toprule
	WMAPE           & MLP    & DetNN  & BNN-10 & BNN-30 & BNN-50 \\
	\midrule
	San Francisco   & 0.0718 & 0.0678 & 0.0658 & 0.0657 & 0.066  \\
	New York City   & 0.0773 & 0.0747 & 0.0763 & 0.0743 & 0.0743 \\
	Boston          & 0.0823 & 0.079  & 0.0778 & 0.077  & 0.0768 \\
	Chicago         & 0.0935 & 0.084  & 0.0807 & 0.0802 & 0.0795 \\
	Washington D.C. & 0.079  & 0.0742 & 0.0758 & 0.0737 & 0.0737 \\
	Seattle         & 0.0822 & 0.0813 & 0.0777 & 0.0772 & 0.077  \\
	Los Angeles     & 0.0792 & 0.0703 & 0.0655 & 0.0647 & 0.065  \\
	Atlanta         & 0.0933 & 0.0877 & 0.0825 & 0.0805 & 0.0813 \\
	\midrule
	Average         & 0.0823 & 0.0774 & 0.0753 & 0.0741 & 0.0742 \\
	\bottomrule
\end{tabular}
  }
\end{table}

The performance of the BNN model with 10, 30 and 50 particle samples, referred to as BNN-10, BNN-30 and BNN-50, is shown in Table~\ref{tbl:wmape}. With only one SVGD sample, a reasonably well maximum a posteriori estimate can be obtained. The sample size in the experiment is chosen arbitrarily as a balance of prediction performance and computational efficiency. The input sequence length is 144 hours, the output sequence length is 6 hours. The weighted mean absolute percentage error (WMAPE) $\sum_{i=1}^N |y_i-\hat{y}_i|/\sum_{i=1}^N |y_i|$, where $y$ and $\hat{y}$ are the true and predicted outcome, is used as the evaluation metric. Table~\ref{tbl:wmape} shows a summary of averaged WMAPE across the 6-hour prediction window. As performance benchmarks, we also show the results of a MLP model and a DetNN model which has the same network structure as the BNN. The hyper-parameters are tuned separately for each model. Averaging across all cities, DetNN achieves 6\% decrease in WMAPE from MLP, and BNN-30 achieves 4\% decrease from DetNN. Bayesian inference of parameters using SVGD further improves the DetNN performance with an additional benefit of quantified prediction variability.

\begin{figure}[ht]
    \vspace{-1.7\baselineskip}
    \centering
    \subfloat{{\includegraphics[width=5.5cm]{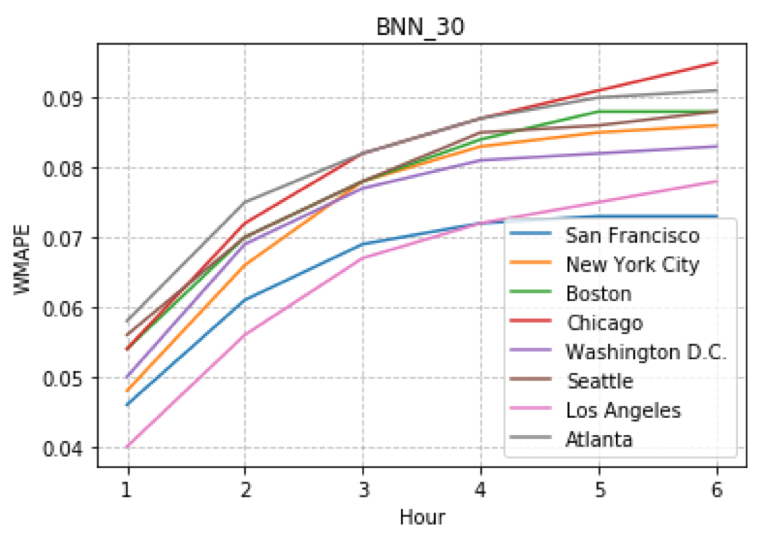} }}
    \qquad
    \subfloat{{\includegraphics[width=5.5cm]{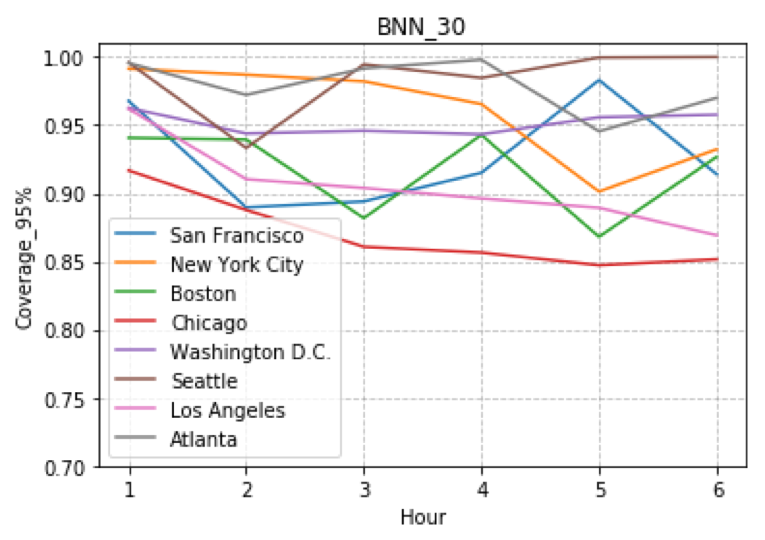} }}
    \caption{Estimated WMAPE (left) and 95\% coverage probability (right) with a 6-hour prediction window.}%
    \label{fig:wmape_cov}
    \vspace{0.7\baselineskip}
\end{figure}

Figure~\ref{fig:wmape_cov} shows the estimated WMAPE and 95\% coverage probability from BNN-30 over a 6-hour prediction window. 95\% coverage probability means the percentage that the true value is within the 95\% credible band. The WMAPE increases when predicting further, but the coverage probability does not necessarily decrease. Even if the point estimation is not good enough, the BNN model could report low confidence by having a high variability around the estimation, thus facilitating better informed supply allocation.

\vspace{-0.75\baselineskip}
\section{Discussion}
\vspace{-0.5\baselineskip}
We have proposed a particular neural network architecture aimed at spatiotemporal modeling with cyclical components applied to the example of estimating demand, which is an important problem in the ridesharing space. We furthermore perform Bayesian inference on the proposed model using a variant of SVGD that gives us promising performance gains.
Our experimental results indicate the advantage of Bayesian estimation using SVGD for our model, which encourages further investigation into the issue of modeling uncertainty for industrial scale problems. There remain interesting research questions to be investigated further. For example, in the future, we will explore different correlation structures to model time series data and investigate the use of more structured prior distributions instead of the independent prior assumption we are currently making.

\bibliographystyle{unsrtnat}
\bibliography{nips_2018}

\section*{Appendix}
\begin{algorithm}[H]
  \begin{algorithmic}[1]
    \Input{data $\mathcal{D}$, step size $e$, target joint likelihood $p(\mathcal{D},\bm{\theta})$}
    \Initialize{n neural networks}
      \For{each batch}
          \State feed forward to compute $
         \log p(\mathcal{D},\bm{\theta}_i)$ for $i=1, \dots, n$.
          \State backpropagate to calculate $\nabla_{\bm{\theta}_i} \log p(\mathcal{D},\bm{\theta}_i)$ for $i=1,\dots, n$.
          \State update $\bm{\theta}_i$ for $i=1, \dots, n$: $\bm{\theta}_i \gets \bm{\theta}_i + \frac{e}{n}\sum_{j=1}^n[k(\bm{\theta}_j,\bm{\theta}_i)\nabla_{\bm{\theta}_j} \log p(\mathcal{D},\bm{\theta}_j) + \nabla_{\bm{\theta}_j}k(\bm{\theta}_j, \bm{\theta}_i)]$
      \EndFor
  \end{algorithmic}
  \caption{BNN via SVGD}
  \label{ag:ag1}
\end{algorithm}

\end{document}